\pdfoutput=1

\documentclass[11pt]{article}

\usepackage[final]{coling}
\usepackage{amssymb}
\usepackage{xcolor}
\usepackage{pifont}    
\usepackage{booktabs}
\usepackage{float}
\usepackage{placeins}
\usepackage{tcolorbox}
\usepackage{multirow}
\usepackage{CJK}
\usepackage{float}
\usepackage{lipsum}
\usepackage{tabularx}
\definecolor{darkgreen}{rgb}{0.0, 0.5, 0.0}
\usepackage{times}
\usepackage{latexsym}

\usepackage[T1]{fontenc}

\usepackage[utf8]{inputenc}

\usepackage{microtype}

\usepackage{inconsolata}

\usepackage{graphicx}

\title{A Multi-Task Role-Playing Agent Capable of Imitating Character Linguistic Styles}

\author{Siyuan Chen$^{1,2}$\thanks{$^\ast$Equal contribution.},\ Qingyi Si$^{3\ast}$,\ Chenxu Yang$^{1,2\ast}$,\ Yunzhi Liang$^{1,2}$ \ Zheng Lin$^{1,2}$\thanks{\ \  Corresponding author: Zheng Lin.}\  \\  {\bf\ Huan Liu$^{1,2}$,\ Weiping Wang$^1$ } \\ 
$^1$Institute of Information Engineering, Chinese Academy of Sciences, Beijing, China \\
$^2$School of Cyber Security, University of Chinese Academy of Sciences, Beijing, China \\
$^3$Huawei Technologies Co., Ltd.\\
  \tt{{\{chensiyuan,yangchenxu,liangyunzhi,linzheng,liuhuan,wangweiping\}@iie.ac.cn,}} \\
  \tt{siqingyi@huawei.com}
  }

\begin{document}
\maketitle
\begin{abstract}
The advent of large language models (LLMs) has significantly propelled the advancement of Role-Playing Agents (RPAs). However, current Role-Playing Agents predominantly focus on mimicking a character's fundamental attributes while neglecting the replication of linguistic style, and they are incapable of effectively replicating characters when performing tasks beyond multi-turn dialogues, which results in generated responses that lack authenticity.
The reason current RPAs lack this capability is due to the nature of existing character datasets, which lack collections of character quotations and are limited to multi-turn dialogue tasks, constraining the RPA's performance across other task domains and failing to mimic a character's linguistic style. To address this gap, we developed a multi-task role-playing dataset named MRstyle, which encompasses a substantial number of real individuals along with their quotations and covers seven different tasks. On this basis, we develop StyleRPA, a Multi-Task Role-Playing Agent  (MRPA) that significantly outperforms recent open-source LLMs and RPAs baselines on 7 tasks including Dialogue, Dictionary, Composition, Story Generation, Product Description, Music Commentary, and Open Question Answering. The code and data will be released.

\end{abstract}

\section{Introduction}

Large Language Models (LLMs)  \cite{openai2024gpt4technicalreport, touvron2023llama2openfoundation} are progressively revolutionizing the paradigm of human-computer interaction. To satisfy fundamental psychological needs of human such as love and belonging \cite{maslow1943theory}, these models are deployed as Role-Playing Agents (RPAs)  \cite{li2023chatharuhirevivinganimecharacter, zhou2023characterglmcustomizingchineseconversational,wang2024rolellmbenchmarkingelicitingenhancing}, engaging in meaningful dialogues with users by skillfully simulating a variety of characters with diverse attributes.

\begin{figure*}[h]
    \centering
    \includegraphics[width=\linewidth]{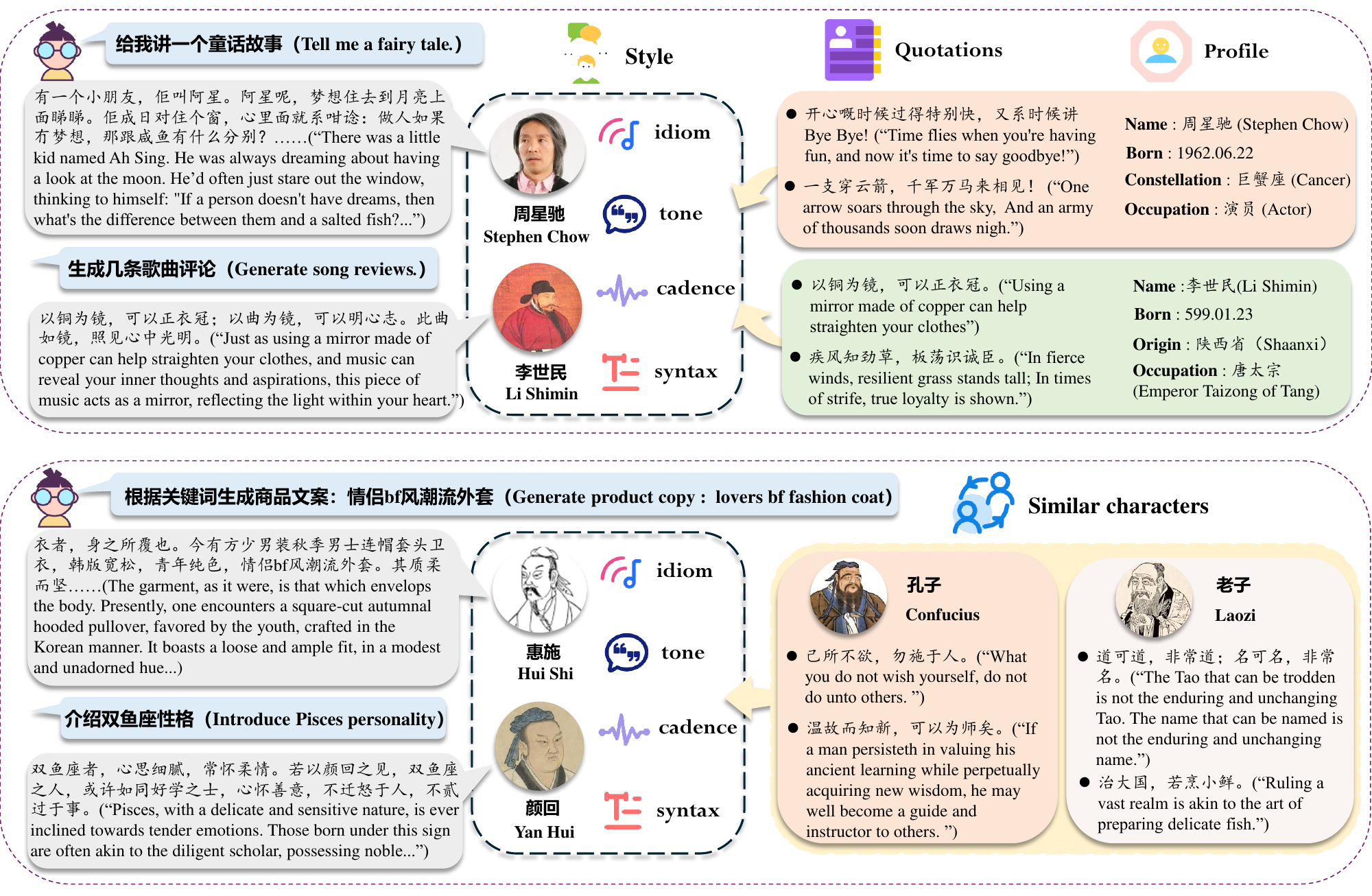}
    \caption{An example of The data in MRstyle. The upper of the chart illustrates the responses generated by a character with quotations, and the lower presents the responses generated by a character without quotations.}
    \label{fig:1}
\end{figure*}

The realization of an ideal RPA presents two significant challenges. First, it should not only maintain consistency in character information but also accurately mimic the character's linguistic style to simulate specific characters more accurately and authentically, as personalized linguistic style is critical in daily interactions and creation  \cite{leech2007style} and lacking linguistic style could reduce user engagement and credibility. Unfortunately, existing datasets\cite{chen2024socialbenchsocialityevaluationroleplaying, zhou2023characterglmcustomizingchineseconversational, gao2023peacokpersonacommonsenseknowledge} for character typically consist of fictional characters and contain only basic information. However, These pieces of information is insufficient to accurately replicate a character's linguistic style. Adding a person's quotations can effectively address this limitation, as such quotations frequently encapsulate their unique linguistic patterns and habits. This limitation hampers the model's ability to replicate how a character would speak in real-life situations, potentially making the user experience feel less credible and realistic. Second, An effective RPA should possess the capability to execute a diverse array of tasks. Nevertheless, the current dataset is limited to dialogue-based tasks, with little application in other contexts such as article or comment generation, which is inadequate for training a robust RPA model. As a result, when the model engages in various types of downstream tasks,  fails to accurately replicate the distinctive linguistic traits of specific characters.

\begin{table*}[h]
\renewcommand{\arraystretch}{1.5}
\small
\begin{tabular}{cccccccccccc}
\hline
\multicolumn{1}{c}{\multirow{2}{*}{Dataset}} & \multicolumn{1}{c}{\multirow{2}{*}{Nums}} & \multicolumn{1}{c}{\multirow{2}{*}{Quotes}} & \multicolumn{7}{c}{Task}& \multicolumn{1}{c}{\multirow{2}{*}{Real}} & \multicolumn{1}{c}{\multirow{2}{*}{Conn}} 
\\ 
\cline{4-10}
\multicolumn{1}{c}{} & \multicolumn{1}{c}{}  & \multicolumn{1}{c}{} & \multicolumn{1}{c}{Dic.} & \multicolumn{1}{c}{Sto.} & \multicolumn{1}{c}{Com.} & \multicolumn{1}{c}{Pro.} & \multicolumn{1}{c}{Ope.} & \multicolumn{1}{c}{Mus.} & \multicolumn{1}{c}{Dia.} & \multicolumn{1}{c}{}  & \multicolumn{1}{c}{}                          
\\ 
\hline
\multicolumn{1}{c}{CharacterEval \cite{tu2024characterevalchinesebenchmarkroleplaying}} & 77 & \textcolor{darkgreen}{\ding{51}} & \textcolor{red}{\ding{55}} & \textcolor{red}{\ding{55}} & \textcolor{red}{\ding{55}} & \textcolor{red}{\ding{55}} & \textcolor{red}{\ding{55}} & \textcolor{red}{\ding{55}} & \textcolor{darkgreen}{\ding{51}} & \textcolor{red}{\ding{55}} & \textcolor{red}{\ding{55}} \\
\multicolumn{1}{c}{PerLTQA \cite{du2024perltqapersonallongtermmemory}}       & 141 & \textcolor{red}{\ding{55}} & \textcolor{red}{\ding{55}} & \textcolor{red}{\ding{55}} & \textcolor{red}{\ding{55}} & \textcolor{red}{\ding{55}} & \textcolor{red}{\ding{55}} & \textcolor{red}{\ding{55}} & \textcolor{darkgreen}{\ding{51}} & \textcolor{red}{\ding{55}} & \textcolor{red}{\ding{55}}  \\
\multicolumn{1}{c}{CharacterGLM \cite{zhou2023characterglmcustomizingchineseconversational}}  & 250  & \textcolor{red}{\ding{55}} & \textcolor{red}{\ding{55}} & \textcolor{red}{\ding{55}} & \textcolor{red}{\ding{55}} & \textcolor{red}{\ding{55}} & \textcolor{red}{\ding{55}} & \textcolor{red}{\ding{55}} & \textcolor{darkgreen}{\ding{51}} & \textcolor{red}{\ding{55}} & \textcolor{red}{\ding{55}} \\
\multicolumn{1}{c}{Socialbench \cite{chen2024socialbenchsocialityevaluationroleplaying}}   & 512  & \textcolor{red}{\ding{55}} & \textcolor{red}{\ding{55}} & \textcolor{red}{\ding{55}} & \textcolor{red}{\ding{55}} & \textcolor{red}{\ding{55}} & \textcolor{red}{\ding{55}} & \textcolor{red}{\ding{55}} & \textcolor{darkgreen}{\ding{51}} & \textcolor{red}{\ding{55}} & \textcolor{darkgreen}{\ding{51}}  \\
\multicolumn{1}{c}{MBTI-1024 \cite{tu2023characterchatlearningconversationalai}}   & 1024 & \textcolor{red}{\ding{55}} & \textcolor{red}{\ding{55}}  & \textcolor{red}{\ding{55}} & \textcolor{red}{\ding{55}} & \textcolor{red}{\ding{55}} & \textcolor{red}{\ding{55}} & \textcolor{red}{\ding{55}} & \textcolor{darkgreen}{\ding{51}} & \textcolor{red}{\ding{55}} & \textcolor{red}{\ding{55}}  \\ 
\hline
\multicolumn{1}{c}{Ours} & \multicolumn{1}{c}{1376} & \multicolumn{1}{c}{\textcolor{darkgreen}{\ding{51}}} & \multicolumn{1}{c}{\textcolor{darkgreen}{\ding{51}}} & \multicolumn{1}{c}{\textcolor{darkgreen}{\ding{51}}} & \multicolumn{1}{c}{\textcolor{darkgreen}{\ding{51}}} & \multicolumn{1}{c}{\textcolor{darkgreen}{\ding{51}}} & \multicolumn{1}{c}{\textcolor{darkgreen}{\ding{51}}}    & \multicolumn{1}{c}{\textcolor{darkgreen}{\ding{51}}} & \multicolumn{1}{c}{\textcolor{darkgreen}{\ding{51}}} & \multicolumn{1}{c}{\textcolor{darkgreen}{\ding{51}}}  & \multicolumn{1}{c}{\textcolor{darkgreen}{\ding{51}}} \\ \hline
\end{tabular}
\caption{\label{table:1}
    Comparison of MRstyle Dataset with Other Datasets. Number of Characters (Nums), character's quotations (Quotes), Real peopl (Real), Connections between characters (Conn), Dictionary (Dic.), Story Generation (Sto.), Composition (Com.), Product Description (Pro.), Open Question Answering (Ope.), Music Commentary (Mus.), and Dialogue (Dia.) .
  }
\end{table*}

To tackle these two challenges, we developed a dataset named MRstyle, which comprises a considerable number of real individuals and their quotations and covers seven different tasks. Specifically, to enhance the understanding of semantic styles (as previously discussed), intuitively, we propose an approach of learning diverse linguistic styles from real individuals and their quotations. Inevitably, some character's quotations remain inaccessible.
Accordingly, we designed two types of scenarios: one aimed at generating responses by emulating characters with available quotations and another focused on emulating those without accessible quotations. For the first scenario, we gathered data on 485 characters with available quotations and developed a individualized framework to summarize their profiles and linguistic styles, prompting LLMs to construct responses that align with the character's linguistic style, as shown in the upper plot of Figure \ref{fig:1}. For the second scenario, we collected 947 characters without quotations and developed a reasoning chain that prompting LLMs to construct the linguistic style of each character based on information from other similar characters' quotations, as shown in the lower plot of Figure \ref{fig:1}. Moreover, the MRstyle dataset comprises seven downstream tasks: Dialogue, Dictionary, Composition, Story Generation, Product Description, Music Commentary, and Open Question Answering. In each task, responses are meticulously crafted to reflect the distinctive linguistic style associated with each character.

Compared to previous datasets, as shown in Table \ref{table:1}, our dataset provides the following advantages: \textit{(1) A substantial collection of authentic characters with associated quotations.} MRstyle is the first character dataset composed of real individuals, systematically organizing their quotations and establishing connections between similar characters. \textit{(2) Inclusion of seven distinct tasks.} MRstyle represents the first multi-task character dataset, ensuring that each character's responses across all tasks are consistent with their unique linguistic style. \textit{(3) The CoT for inferring linguistic style from character quotations.} We have developed prompts and CoT reasoning methods that infer characters' linguistic styles based on their quotations.

Based on the curated dataset, we develop a multi-task RPA named StyleRPA. StyleRPA is capable of modeling the linguistic style of a character based on diverse characters, generating responses that are more aligned with the character's distinctive language patterns. We conducted a comparative analysis of StyleRPA against several open-source LLMs and closed-source GPT-4o. The experimental results show that StyleRPA surpasses the LLM baseline across seven distinct tasks and is even competitive with the powerful GPT-4o in some tasks. We anticipate that this research will establish a solid foundation for future inquiries into large-scale model-based role-playing challenges and foster advancements in more versatile RPAs.

We conclude our contributions as follows:

(1) We create a multi-task character dataset MRstyle, which collects a large number of real people and their quotes and includes seven tasks.

(2) We developed MRstype-instruct character tuning data, which summarizes the character's language style and analyzes five different real-life scenarios, can be better used for model fine-tuning.

(3)  We develop a multi-task RPA named StyleRPA, which outperform open-source LLMs in seven tasks, even exceeding GPT-4o 5\% in the scenario of mimicking characters with quotations.

\section{Related Work}

\subsection{Personalized Dialogue}
Personalized dialogue systems, which generate responses in accordance with specific personas, have garnered significant attention in recent research. The focus was on maintaining consistency between persona attributes and the generated responses    \cite{cheng-etal-2023-pal,chen2023learningmemorizeentailmentdiscourse}. Persona-based dialogue datasets  \cite{DBLP:journals/corr/abs-1801-07243, dinan2019secondconversationalintelligencechallenge, xu2022longtimeseeopendomain} are constructed through interactions between two human speakers, each providing approximately five sentences to describe their persona traits, which are then reflected in the subsequent dialogue. Although these studies explored character-related issues in dialogue, the personas used are often limited to basic personal details such as name, age, and experience  \cite{jang2022customizedconversationcustomizedconversation, gao2023peacokpersonacommonsenseknowledge}. Responses generated based on these personas typically only convey basic character information and fail to reflect distinct linguistic styles, resulting in replies that resemble generic answers and lack personalized characteristics.

\subsection{Role-Playing Agents}
Recent research advanced the key capabilities of large-scale language models (LLMs) within the field of RPAs \cite{10.1145/3586183.3606763, shanahan2023roleplaylargelanguagemodels,wang2024incharacterevaluatingpersonalityfidelity} . These agents can assume various roles by effectively simulating role data \cite{tu2024characterevalchinesebenchmarkroleplaying, wang2024rolellmbenchmarkingelicitingenhancing}, which is derived from prompt contexts, training sets, or external databases. Existing studies have enhanced RPAs' performance by setting role descriptions as system prompts \cite{li2023chatharuhirevivinganimecharacter, tu2023characterchatlearningconversationalai}. Furthermore, specialized role training corpora can be provided and fine-tuned based on open-source foundational models \cite{Shen2023RoleEvalAB}. To further enhance the capabilities of RPAs, researchers have integrated retrieval-augmented methods; for example, \citet{salemi2024lamplargelanguagemodels} combined role-playing model training with retrieval information to improve agent performance in role-playing tasks. The combination of these approaches demonstrates that high-quality role data is crucial for the performance of RPAs.

\section{MRstyle}
The process of collecting MRstyle is primarily divided into two phases: Character Data Collection and Task Data Construction, as detailed in Sections 3.1 and 3.2. Character Data Collection is aimed at gathering character information and quotations, while Task Data Construction focuses on building downstream task data.
\subsection{Character Data Collection}
The MRstyle dataset is primarily composed of two categories of characters. The first category includes characters with quotations, referred to as seed characters, from which RPAs can infer the linguistic style based on these quotations. For this latter group, we establish associations with similar seed characters to enable RPAs to deduce the linguistic style of sprout characters using the quotations from their corresponding seed characters and generate appropriate responses. Our character collection process is outlined below:

\paragraph{Collect characters' information.}
First, we extracted a substantial amount of Chinese character data from Wikipedia and eliminated any tags associated with political or commercial activities to ensure that the dataset was free from biased or propagandistic content. We standardized the information and further removed entries with insufficient content to guarantee that each character possessed adequate information. This process ensured that all characters were supported by sufficiently rich data.

\paragraph{Collect characters' quotations.}
Subsequently, we extracted a selection of quotations from Wikipedia\footnote{https://zh.wikipedia.org/wiki/Wikipedia} and conducted a rigorous screening process, eliminating quotations attributed to politically affiliated figures, those with negative social impacts, as well as quotations reflecting political stances or personal opinions to ensure the neutrality and objectivity of the dataset. Acknowledging that the quotations available on Wikipedia were insufficiently comprehensive and lacked certain classic sayings, we supplemented these figures' quotations by sourcing additional material from the internet. For contemporary figures, we gathered their public speeches and incorporated them into the dataset; for historical figures, we selected notable excerpts from their classical writings to ensure that each figure had no fewer than three quotations, thereby guaranteeing both completeness and breadth of content.

Ultimately, we acquired 485 seed characters accompanied by quotations, each with 3 to 10 quotations, and 947 budding characters without quotations. Each budding character was associated with a corresponding similar seed character. All these character information is organized in a key-value format, and further details can be found in the appendix \ref{sec:MRstyle}.

\subsection{Task Data Construction}
In the previous section, we gathered a wealth of characters and related information. In this section, we will use this character information to prompt GPT-4 to generate instruction-response pairs tailored to different tasks. This is crucial for ensuring that the model maintains a consistent character linguistic style when handling various tasks. To enhance the model's reasoning and generalization abilities, we also developed specific CoT for each character, guiding the model to imitate the character's linguistic style rather than merely reproducing quotations.


\paragraph{Collecting multi-task data.}
We selected two open-source Chinese datasets, NaturalConv\footnote{https://github.com/naturalconv/NaturalConvDataSet} and YeungNLP/firefly-train-1.1M\footnote{https://huggingface.co/datasets/YeungNLP/firefly-train-1.1M}, to obtain original data for different tasks. NaturalConv is a dataset of daily conversations used to construct character responses in daily dialogue, and YeungNLP/firefly-train-1.1M is a multi-task dataset utilized for generating character responses across six additional tasks: Dictionary, StoryGeneration, Compostion, ProductDesc,  OpenQA, MusicComment. For each task, we matched 3-4 pieces of data to each character, directly reusing the instructions from the task data we collected. Then, based on their original responses, we constructed new responses that align with the character's linguistic style.    

\paragraph{Generate responses by posterior information.}
We observed that using prompt engineering to make GPT-4o directly generate responses imitating a character results in the model simultaneously focusing on content generation, the generated content tends to lack a distinctive character-specific linguistic style and appears more like a generic response. To mitigate this issue, we introduce a posterior information data construction method. Initially, we provide the model with a reference response for original task and make it to contemplate how to produce a similar reply using the imitated character's linguistic style—termed posterior information (For example, the xx in a character's quote can be effectively integrated with the xx in the story). Following this, we instructed GPT-4o to emulate the character based on their quotes alongside the posterior information to generate responses. We sampled and verified the generated results, and found that the response generation effect based on posterior information was better than directly imitating the character's quotations to generate responses. More detailed information is provided in the appendix \ref{sec:Posterior}.

\paragraph{Develop CoT.}
We then compiled the collected character data and generated response data, developing specific CoT for each character. For seed characters with quotations, our CoT summarized the character's information and the linguistic characteristics of their quotations, guiding the model to infer and apply the character's linguistic style to the generated responses. For budding characters without quotations, our CoT summarized the linguistic styles of similar seed characters, inferred the linguistic style of the budding character, and guided the generation of character-appropriate responses. 


To further ensure the quality of the data, we also selected three experienced annotators to check the annotation results and select low-quality responses for re-generation. These data, including various real-life characters, real quotes, varying tasks with different linguistic-style responses, can be used to instruction-tuning our custom models, enabling them to imitate the linguistic style of characters to complete different tasks.
\subsection{Dataset Analysis}
The MRstyle framework offers the following advantages: \textit{(1) A substantial number of real characters with authentic quotations.} It includes 485 characters with authentic quotations and 947 characters without quotations. Comprehensive information has been meticulously compiled for each character, ensuring that this information is publicly available and transparent, without involving personal opinions, biases, or commercial actions. For the first time, we systematically organized characters' quotations and established connections between similar characters. \textit{(2) Multiple task inclusion.} The framework encompasses seven distinct tasks, amounting to a total of 31,117 data entries. Each character has responses under each task, and we ensure that their responses align with the linguistic style of the character. \textit{(3) The CoT for inferring linguistic style from character quotations.} Personalized Prompts and CoT were designed for each character under every task, guiding the model to infer the character's linguistic style, thus generating responses that more closely match the character’s manner of speaking.  Further details can be found in the appendix \ref{sec:static}.


Our train-test split strategy focuses on two dimensions: (1) Instruction-based splitting, used to assess the model's generalization capability on new instructions. For this, we selected 3,000 test samples across seven tasks; (2) Character-based splitting, used to evaluate the model's ability to generalize to new characters, which were not encountered by the model during the training. We selected 10 new characters for testing, and the model had not encountered these characters during training.

\section{StyleRPA}
After constructing the MRstyle dataset, we endeavor to fully leverage this data to promote model's ability to imitate the linguistic style of the given characters. Inspired by the multi-task training paradigm adopted by previous RPAs \cite{du2024perltqapersonallongtermmemory, lu2024largelanguagemodelssuperpositions}, we devise a two-scenario instruction fine-tuning framework based on potential real-world user inputs and choose Qwen2-7B-Base as the backbone to develop a multi-tasks RPA named StyleRPA. Our model can adapt to character data with varying amounts of information and is able to better reason and mimic the given character's linguistic style.

\subsection{MRstyle-instruct for Scenarios}
To address the various characteristics that need to be simulated in RPAs, which can lead to significant differences in user input, we have simulated the following scenarios to enhance the model's generalization capability as much as possible.
  
\paragraph{Seed Characters.}
In real-world scenarios, quotations from certain individuals, such as well-known actors or writers, are relatively easy to obtain. These individuals are referred to as "seed characters" in our MRstyle dataset. We designed two types of input formats for instruction fine-tuning data for these characters: 1) Providing both character information and quotations. In this case, the model infers the character's linguistic style based on the given quotations. 2) Providing only the information of the widely acknowledged character. The model weights contain quotes from these characters, We expect the model to imitate the target character's linguistic style solely based on their identity, by the linguistic style into the model's weights. Correspondingly, we also designed two output formats: 1) Using CoT reasoning, where the model infers the linguistic style from the quotations and generates a response, applicable to Input 1. 2) Directly generating responses that imitate the character, applicable to both Input 1 and Input 2. Our instruction fine-tuning data covers all combinations of inputs and outputs, with the ratio of Input 1 to Input 2 being 4:1. Detailed data is presented in the appendix \ref{sec:MRstyle}.

\paragraph{Budding Characters.}
Considering that the quotations of some characters are difficult to obtain, we are often only able to collect their information. These characters are referred to as "budding characters" in our MRstyle dataset. Due to the lack of quotation data, we need to retrieve similar characters and infer the linguistic style of the budding characters based on their quotations. For such characters, we have designed three input forms to handle potential scenarios: 1) Providing both the target character's information and quotations from similar characters. In this case, the model infers the linguistic style of the target character from the quotations of similar characters and generates appropriate responses. 2) Providing only the target character’s information without any reference characters. In this scenario, we associate the target character with similar seed characters in MRstyle. Since seed characters have available quotations, the model can infer the linguistic style of the target character and generate responses. 3) Providing well-known characters without any additional background information. We expect the model to internalize the character's linguistic style directly into its weights and generate responses consistent with that character’s style, without relying on external knowledge. Correspondingly, we have also designed two output forms: 1) Using CoT reasoning, where the model deduces the target character’s linguistic style based on similar characters' quotations, applicable to Input 1 and Input 2. 2) Directly generating responses in the style of the target character, applicable to Input 1, Input 2, and Input 3. Our instruction fine-tuning dataset covers all input-output combinations, with the ratios of Input 1, Input 2, and Input 3 being 2:2:1.

\subsection{Model Training}
We organized the characters in the MRstyle dataset according to the two aforementioned scenarios, constructing the Pstyle-instruct to align with our input-output settings. We mixed Pstyle-instruct with seven general task datasets, for model fine-tuning. Although there are already numerous instruction-tuning datasets for role-playing, these datasets fail to sufficiently model character-specific linguistic styles, limiting their ability to accurately imitate particular personas. Pstyle-instruct aims to fill this gap, enabling the model to emulate different characters' linguistic styles across various scenarios. During training, we also introduced some general fine-tuning data to enhance the model's generalization capabilities, maintaining a ratio of 6:5 between general and ours data.

\begin{table*}[h]
  \centering
  \setlength{\tabcolsep}{4pt}
  \begin{tabular}{ccccccccc}
    \hline
    \textbf{Model}           & \textbf{Dic.} & \textbf{Sto.} & \textbf{Com.} & \textbf{Pro.} & \textbf{Ope.} & \textbf{Mus.} & \textbf{Dia.} & \textbf{Avg.}  \\
    \hline
      GPT-4o       &45.1&42.2&47.4&45.5&48.7&49.5&\textbf{56.0}&47.8   \\
    \hline
      ChatGLM3-6B    &3.6&9.0&10.8&10.9&12.2&7.8&13.8&9.7  \\
      Yi-6B-Chat       &4.6&11.3&12.2&14.6&16.0&8.5&17.7&12.1  \\
      ChatPLUG &2.6&3.7&6.0&8.9&10.1&9.1&12.9&7.6     \\
    CharacterGLM-6B   &6.9&12.6&14.9&12.0&9.8&10.9&19.3&12.3\\
        \hline
    StyleRPA  &\textbf{54.9}&\textbf{57.8}&\textbf{52.6}&\textbf{54.5}&\textbf{51.3}&\textbf{50.5}&44.0&\textbf{52.2}          \\
        \hline
  \end{tabular}
  \caption{\label{table:3}
   The winning rate of models compared to GPT-4o on the new  instructions test set with 7 task. (\textit{The winning rate of GPT-4o was obtained by comparing it with StyleRPA})
  }
  
\end{table*}
\section{Experiments}
\subsection{Experimental Setup}
\paragraph{Baseline.} We consider baselines of three genres:(1) Open-source Large Language Models(LLMs) including ChatGLM3-6B \cite{glm2024chatglmfamilylargelanguage}, Yi-6B-Chat \cite{ai2024yiopenfoundationmodels}. (2) Open-source Character Large Language Model(CLLMs) including 
CharacterGLM-6B \cite{zhou2023characterglmcustomizingchineseconversational}, ChatPLUG \cite{tian2023chatplugopendomaingenerativedialogue}. (3) GPT-4o with role mimicry capability\cite{wang2024rolellmbenchmarkingelicitingenhancing}. More detailed information is provided in the appendix \ref{sec:prompt}.
\paragraph{Evaluation metrics.} The similarity of linguistic style, such as speaking rhythm, tone, and intonation, is challenging to quantify using the overlap between predicted and true values. Therefore, we did not employ automated metrics to evaluate our model. To more accurately assess the quality of the generated content's linguistic style, we referred to the method used in AlpacaEval \cite{dubois2024lengthcontrolledalpacaevalsimpleway}. We utilized GPT as the evaluator, as its reliability has been demonstrated \cite{fu2023gptscoreevaluatedesire, Gilardi_2023}. Considering the cost of GPT-4, we opted for a lighter model, GPT-4o, with comparable performance for evaluation. Our evaluator’s prompts were slightly modified from the AlpacaEval prompts, focusing on the similarity of the character's linguistic style and the accuracy of the generated responses. During evaluation, the similarity between the model-generated responses and the actual linguistic style of these characters' quotations will be compared, as detailed in the appendix \ref{sec:prompt}. These prompts facilitate sample comparison and ranking, leading to a winning rate evaluation. Specifically, the performance of each model was compared against GPT-4o, which is capable of mimicking specific characters' linguistic styles, and the winning rate of each model was recorded. The winning rate of GPT-4o was obtained by comparing it with StyleRPA. Such a comparison provides a clearer reflection of the performance differences across the seven tasks.

\subsection{Experiment and Analysis}
\paragraph{Performance of new instructions.}
Table \ref{table:3} presents the GPT evaluation performance of the models’ ability to mimic a character's linguistic style across seven tasks. These characters had appeared during the model's training phase, but the data for each task were entirely new. We found that StyleRPA outperformed all baseline models on six tasks. Although it was slightly inferior to GPT-4o in the multi-turn dialogue task due to limitations in model size and corpus volume, it still surpassed the existing baseline models. In contrast, previous open-source large language models and role-playing models showed inadequate performance in mimicking specific characters' linguistic styles. 

\paragraph{Performance of unseen characters. }
Table \ref{table:4} presents the GPT evaluation scores for mimicking the linguistic style of characters on seven tasks, where the model does not encounter these characters or their quotations during training. Given that it is difficult to verify whether any open-source large models or role-playing models have seen these characters during training, we only compared our model with GPT-4o in this section. Even characters not in the training data, StyleRPA outperformed GPT-4o in three tasks, and on other tasks, our model's performance was generally close to that of GPT-4o. Considering that GPT-4o has access to these characters' quotations during training, while this information is entirely new to our model, this demonstrates that the inclusion of character quotations and CoT reasoning significantly enhances the model's ability to mimic the linguistic style of previously unseen characters.

\paragraph{Performance of seed character and budding character.}
We utilized GPT to evaluate the model's performance in mimicking seed characters and budding characters. The linguistic style of this group of characters is inferred from the quotations of related characters. As shown in Table \ref{fig:2}, For the seed character, StyleRPA outperformed GPT-4o on six tasks. For the budding character, StyleRPA outperformed GPT-4o on only 4 tasks. In conclusion, StyleRPA performs better in imitating seed characters, and it is more challenging to imitate budding characters, with a slight decrease, but it is generally better than GPT-4o. The latter indicates that even in the absence of quotations from the target character, our model can still accurately infer and imitate the character’s linguistic style. 

  

\begin{table*}[h]
  \centering
  \setlength{\tabcolsep}{4pt}
  \begin{tabular}{ccccccccc}
    \hline
    \textbf{Model}           & \textbf{Dic.} & \textbf{Sto.} & \textbf{Com.} & \textbf{Pro.} & \textbf{Ope.} & \textbf{Mus.} & \textbf{Dia.} & \textbf{avg.}  \\
    \hline
      GPT-4o       &48.7&45.9&\textbf{51.8}&47.5&\textbf{53.0}&\textbf{57.7}&\textbf{57.9}&\textbf{51.8}   \\
    \hline
    StyleRPA  &\textbf{51.3}&\textbf{54.1}&48.2&\textbf{52.5}&47.0&42.3&42.1&48.2         \\
        \hline
  \end{tabular}
  \caption{\label{table:4}
    The winning rate of models compared to GPT-4o on the new  characters test set with 7 task.
  }
  
\end{table*}

\subsection{Ablation Study}
Considering the cost issue of GPT's API, we randomly selected 200 samples from the test set for evaluation during the ablation experiments. These 200 samples covered all 7 tasks. After averaging the evaluation results, we compared them with the results of StyleRPA to reflect the overall importance of each module. We denote s1 as the performance of seed characters, s2 as the performance of unseen characters, and s3 as the performance of budding characters.

\paragraph{Effect of CoT.}
Table \ref{table:6} illustrates the performance of StyleRPA when trained without the use of Chain-of-Thought (CoT) reasoning. In the absence of CoT, the model's performance deteriorated across all three scenarios, with particularly pronounced declines observed in the second and third scenarios. This decline can be attributed to the fact that these scenarios necessitate the model's ability to infer character linguistic styles based on new quotations and similar characters. Omitting CoT compromises its reasoning capabilities, ultimately impacting the quality of generated outputs.

  

\paragraph{Effect of quotations. }
The incorporation of character quotations markedly improves the ability of RPAs to emulate the linguistic styles of characters, as shown in Table \ref{table:7}. When there is no character quotations as a reference, the model can only rely on the knowledge acquired during training to reason and match the linguistic style of the given character during the generation process. This makes it difficult for the model to fully mimic the specific characteristics of a character. From the results, it can be seen that character quote data is very important for improving the imitation ability of RPAs.

  

\textbf{Effect of Similar Characters. }
When given characters do not have a corpus of quotations, providing the model with quotations from similar characters can also enable the model to infer the character's linguistic style and generate responses, as shown in \ref{table:8}.  In the absence of similar characters, the model is limited to deducing the character's linguistic style solely based on its information, impairing its capacity to replicate the given character.  The experimental results prove that it is important to provide similar characters with quotations when characters do not have a corpus of quotations.

  

\begin{table}[h]
  \centering
  \begin{minipage}{0.32\textwidth}
    \centering
    \setlength{\tabcolsep}{4pt}
    \begin{tabular}{cccc}
      \hline
      \textbf{}           & \textbf{S1} & \textbf{S2} & \textbf{S3}   \\
      \hline
      w/o CoT       &41.5&31.0&32.5   \\
      \hline
      StyleRPA  &\textbf{53.5}&\textbf{48.2}&\textbf{50.9}       \\
      \hline
    \end{tabular}
    \captionsetup{width=1.5\textwidth} 
    \caption{\label{table:6} The winning rate of StyleRPA and StyleRPA without CoT compared to GPT-4o.}
  \end{minipage}
  \hfill
  \begin{minipage}{0.32\textwidth}
    \centering
    \setlength{\tabcolsep}{4pt}
    \begin{tabular}{cc}
      \hline
      \textbf{}           & \textbf{S1} \\
      \hline
      w/o Quotations       &33.6   \\
      \hline
      StyleRPA  &\textbf{53.5}      \\
      \hline
    \end{tabular}
    \captionsetup{width=1.5\textwidth} 
    \caption{\label{table:7} For the seed characters, the winning rate of StyleRPA and StyleRPA without quotations compared to GPT-4o.}
  \end{minipage}
  \hfill
  \begin{minipage}{0.32\textwidth}
    \centering

    \setlength{\tabcolsep}{4pt}
    \begin{tabular}{cc}
      \hline
      \textbf{}           & \textbf{S3} \\
      \hline
      w/o Similar Characters       &31.7   \\
      \hline
      StyleRPA  &\textbf{50.9}      \\
      \hline
    \end{tabular}
    \captionsetup{width=1.5\textwidth} 
    \caption{\label{table:8} For the budding characters, the winning rate of StyleRPA and StyleRPA without similar characters to GPT-4o.}
  \end{minipage}
\end{table}











\section{Conclusion}
The paper represents a pioneering effort to investigate the imitation of character linguistic styles by large models across various tasks. We introduce a multi-task role-playing dataset, MRstyle, which encompasses a substantial collection of real-life characters along with their quotations, spanning seven distinct tasks. Furthermore, We developed a fine-tuning dataset for MRstyle-instruct through the integration of diverse scenarios. On the basis of MRstyle-instruct, we empower Qwen2 to become a multi-task Role-Play Agent StyleRPA, which proficient in emulating diverse character linguistic styles. Experimental results show that StyleRPA significantly outperforms existing RPAs on multiple tasks, and is even on par with the powerful GPT-4o. In conclusion, the contributions of this paper lie at promoting the research on Role-Play Agent understanding from the task, dataset and model perspectives.

\section*{Limitations}
Although this work provides a comprehensive exploration of Role-Playing Agent Capable of Imitating Character Linguistic Styles on multi-tasks for the first time, there are still some limitations that can be left for further research. Firstly, our dataset only contains Chinese data, lacking data in some other languages. Secondly, our model has not exceeded GPT-4o in the multi-turn dialogue scenario, and we hope to narrow this gap in the future. Finally, our model's ability to imitate unseen characters needs to be improved. With the emergence of more powerful LLMs, we look forward to developing more realistic and powerful RPAs in the future.

\section*{Ethical Considerations}
The proposed MRstyle dataset is constructed based on the academic datasets like NaturalConv and YeungNLP/firefly-train-1.1M. Furthermore, all of our character information and quotes are sourced from Wikipedia and Wikiquote, which are free and open datasets for research use with licenses like MIT License1 or CC-BY-SA4.0 License 2 . We eliminate any characters associated with political or commercial activities to ensure that the dataset was free from biased or propagandistic content and quotations attributed to politically affiliated figures, those with negative social impacts, as well as quotations reflecting political stances or personal opinions to ensure the neutrality and objectivity of the dataset. The resulting dataset MRstyle is also a free and open resource for the community to study the RPA. Thus, the authors foresee no ethical concerns with the research in this paper.
\bibliography{custom}

\appendix
\begin{figure*}[h]
    \centering
    \includegraphics[width=\linewidth]{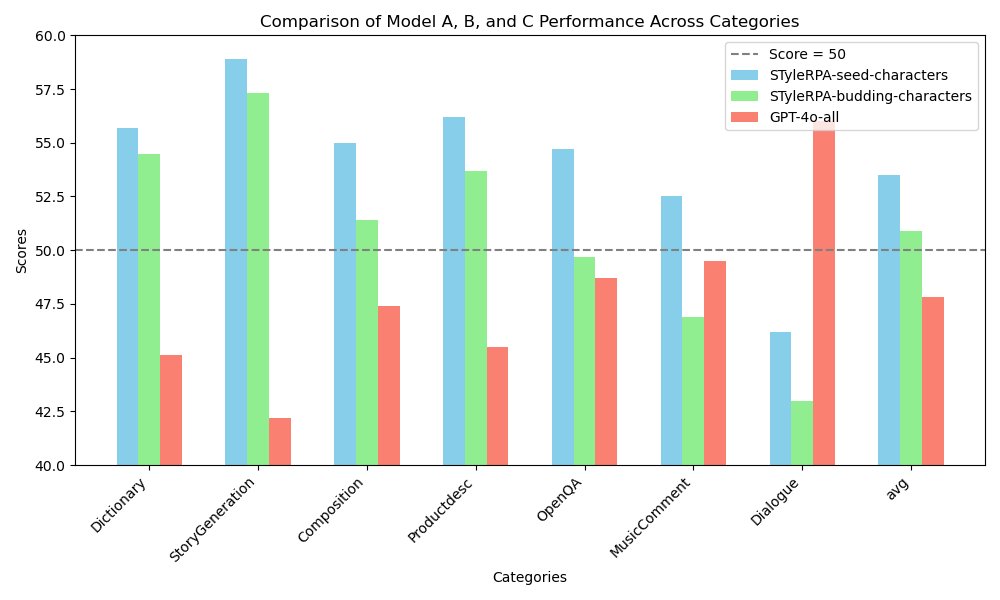}
    \caption{The comparison results of STyleRPA-seed-characters, STyleRPA-budding-characters, and GPT-4o-all.}
    \label{fig:2}
\end{figure*}
\section{Dataset Statistics}
\label{sec:static}
Table \ref{data:static} shows the statistical data of our MRstyle dataset. We have statistically analyzed the number of characters, quotes, and downstream tasks in our dataset.
\begin{table}[htbp]
\centering
\begin{tabular}{lrr}
\toprule
\textbf{Characters} & & \\ 
\# Seed Characters  & 485 \\
\# Budding Characters & 947 \\ 
\# Quotations & 3208 \\
\midrule
\textbf{Tasks} & & \\ 
\# Composition & 3469 \\ 
\# Dictionary & 3033 \\ 
\# MusicComment & 3276 \\ 
\# OpenQA & 3573 \\ 
\# ProductDesc & 3604 \\ 
\# StoryGeneration & 3306 \\ 
\# Dialogue & 11956 \\ 
\bottomrule
\end{tabular}
\caption{Dataset Statistics}
\label{data:static}
\end{table}
\section{Performance of Seed Character and Budding Character.}
Figure \ref{fig:2} shows the comparison between the results achieved by StyleRPA in imitating the seed character and the budding character, and the GPT-4o.

\section{Example from MRstyle Dataset.}
\label{sec:MRstyle}

The seed character datas in the MRstyle dataset are shown in Table \ref{character:1} and \ref{character:2} and the budding characters in Table \ref{character:3}. The data related to the downstream tasks is illustrated in Tables \ref{character:4} and \ref{character:5}.

\begin{table*}[]
\centering
\scalebox{.82}{
\begin{tabular}{c|p{15cm}}
\hline
Character & Examples \\
\hline

Name & \begin{CJK}{UTF8}{gkai}孟子 \end{CJK} \\ 
\cline{2-2}
Information & \begin{CJK}{UTF8}{gkai}
中文名: 孟子\newline
字: 子车/子舆 \newline
所处时代: 战国时期 \newline
民族族群: 华夏族 \newline
出生地: 邹国（今山东邹城） \newline
出生日期: 约公元前372年，一说约前385年 \newline
逝世日期: 公元前289年，一说约前304年 \newline
主要作品: 《孟子》  \newline
尊称: 亚圣
\end{CJK} \\
\cline{2-2}
Quotations & \begin{CJK}{UTF8}{gkai}
仁者无敌。 \newline
保民而王，莫之能御也。 \newline
君子之于禽兽也，见其生，不忍见其死；闻其声，不忍食其肉。是以君子远庖厨也。 \newline
老吾老，以及人之老；幼吾幼，以及人之幼。 \newline
权，然后知轻重；度，然后知长短。物皆然，心为甚。 \newline
我善养吾浩然之气。 \newline 
以力服人者，非心服也，力不赡也；以德服人者，中心悦而诚服也。 \newline
无恻隐之心，非人也；无羞恶之心，非人也；无辞让之心，非人也；无是非之心，非人也。 \newline
恻隐之心，仁之端也；羞恶之心，义之端也；辞让之心，礼之端也；是非之心，智之端也。 \newline
天时不如地利，地利不如人和。
\end{CJK}
\\
\hline
Name & \begin{CJK}{UTF8}{gkai}曹操 \end{CJK} \\ 
\cline{2-2}
Information & \begin{CJK}{UTF8}{gkai}
中文名: 曹操 \newline
别名: 曹吉利、曹孟德、阿瞒（或称曹瞒）\newline 
字: 孟德 \newline
谥号: 武王，武皇帝（曹丕追谥） \newline
封号: 武平侯、魏公、魏王 \newline
庙号: 太祖（曹丕追尊） \newline
所处时代: 东汉末年 \newline
民族族群: 汉族 \newline
出生地: 沛国谯县（今安徽亳州） \newline
出生日期: 155年 \newline
逝世日期: 220年3月15日 \newline
逝世地: 洛阳 \newline
陵墓: 高陵 \newline
继任: 魏王曹丕（魏高祖文皇帝） \newline
主要作品: 观沧海、龟虽寿、述志令、蒿里行、孟德新书、孙子略解 
\end{CJK} 
\\
\cline{2-2}
Quotations & \begin{CJK}{UTF8}{gkai}
有其志必成其事，盖烈士之所徇也。  \newline
生子当如孙仲谋  \newline
老骥伏枥，志在千里；烈士暮年，壮心不已。  \newline
宁我负人，毋人负我。  \newline
人情见利而进，遭害而退。民可与乐成，不可与虑始。  \newline
对酒当歌，人生几何？譬如朝露，去日苦多。  \newline
慨当以慷，忧思难忘。何以解忧，唯有杜康。  \newline
山不厌高，水不厌深，周公吐哺，天下归心。  \newline
任天下之智力，以道御之，无所不可。  \newline
夫英雄者，胸怀大志，腹有良谋，有包藏宇宙之机，吞吐天地之志也。  \newline
汝妻子吾自养之，汝勿虑也。  \newline
今天下英雄，惟使君与操耳！  \newline
设使天下无有孤，不知当有几人称王，几人称帝。或者人见孤彊盛，又性不信天命之事，恐私心相评，言有不逊之志，妄相忖度，每用耿耿。
\end{CJK}
\\
\hline

\end{tabular}
}
\caption{\label{character:1}
Examples of the seed character in MRstyle. 
} 

\end{table*}

\begin{table*}[]
\centering
\scalebox{.82}{
\begin{tabular}{c|p{15cm}}
\hline
Character & Examples \\
\hline

Name & \begin{CJK}{UTF8}{gkai}赵本山 \end{CJK} \\ 
\cline{2-2}
Information & \begin{CJK}{UTF8}{gkai}
中文名: 赵本山 \newline
性别: 男 \newline 
民族: 汉族  \newline
国籍: 中国  \newline
出生地: 辽宁省铁岭市开原市  \newline
出生日期: 1957年10月2日  \newline
星座: 天秤座  \newline
血型: B型 \newline
身高: 175cm \newline
体重: 70kg \newline
职业: 演员、导演、编剧 \newline
别名: 本山大叔
\end{CJK} \\
\cline{2-2}
Quotations & \begin{CJK}{UTF8}{gkai}
好的东西不费劲 费劲的东西都不是好的。 \newline
上顿陪，下顿陪，终于陪出胃下垂。先用盅，再用杯，用完小嘴对瓶吹。 \newline
你让瞎子去南极，他根本找不着北；你让脑血栓去下叉，他根本勾不着腿；你让大马猴穿旗袍，它根本看不出美；你让潘长江去吻赵海霞，他根本够不着嘴。 \newline
那次卖拐把他忽悠瘸了，那次卖车把他忽悠捏了，今天在十分钟之内我要不把他摆平，我就没法跟你们俩当教师爷了！ \newline
谁说我的脸长得像鞋拔子，这是典型的猪腰子脸。\newline 
改革春风吹满地，中国人民真争气，齐心合力跨世纪，一场大水没咋地。\newline 
我允许你走进我的世界，但绝不允许你在我的世界里走来走去。 \newline
恋是个很强悍的字。它的上半部取自“变态”的“变”下半部取自“变态”的“态” \newline
如果中了一千万，我就去买30套房子租给别人，每天都去收一次房租。哇咔咔~~充实！ \newline
人生在世屈指算，最多三万六千天；家有房屋千万所，睡觉就需三尺宽；说人好比盆中鲜花；生活就是一团乱麻！
\end{CJK}
\\
\hline
Name & \begin{CJK}{UTF8}{gkai}雷军 \end{CJK} \\ 
\cline{2-2}
Information & \begin{CJK}{UTF8}{gkai}
中文名: 雷军  \newline
外文名: LeiJun  \newline
国籍: 中国  \newline
民族: 汉族  \newline
出生日期: 1969年12月16日  \newline
毕业院校: 武汉大学  \newline
职务: 董事长、首席执行官（CEO）  \newline
职业: 企业家  \newline
出生地: 湖北省仙桃市剅河镇赵湾村四组
\end{CJK} 
\\
\cline{2-2}
Quotations & \begin{CJK}{UTF8}{gkai}
天下武功，唯快不破。互联网竞争的利器就是快。\newline
当你的前途一片黑暗时，你该意识到是你在发光。\newline
不服跑个分。 \newline
让每个人都能享受科技的乐趣。 \newline
Are you OK？ \newline
找人不是要三顾茅庐，找人要三十次顾茅庐。 \newline
成功往往不是规划出来的，危机是你想不到的机会。 \newline
优秀的公司赚取利润，伟大的公司赢得人心。 \newline
小米三大铁律：技术为本、性价比为纲、做最酷的产品。\newline
相信自己，一往无前。
\end{CJK}
\\
\hline

\end{tabular}
}
\caption{\label{character:2}
Examples of the seed character in MRstyle. 
} 

\end{table*}

\begin{table*}[]
\centering
\scalebox{.82}{
\begin{tabular}{c|p{15cm}}
\hline
Character & Examples \\
\hline
Name & \begin{CJK}{UTF8}{gkai}张大千 \end{CJK} \\ 
\cline{2-2}
Information & \begin{CJK}{UTF8}{gkai}
中文名: 张大千  \newline
别名: 大千居士、下里港人  \newline
国籍: 中国  \newline
民族: 汉族  \newline 
出生地: 四川省内江市市中区城郊安良里  \newline
出生日期: 1899年5月10日  \newline
逝世日期: 1983年4月2日  \newline
毕业院校: 京都公平学校  \newline
职业: 画家  \newline
代表作品: 《荷花图》《爱痕湖》《长江万里图》《秋曦图》等  \newline
字: 季爰 \newline
号: 大千、大千居士、下里港人  \newline
信仰: 佛教 \newline
原名: 正权
\end{CJK} \\
\cline{2-2}
Similar Character & \begin{CJK}{UTF8}{gkai}
中文名: 齐白石  \newline
国籍: 中国  \newline
民族: 汉族  \newline
出生地: 湖南湘潭  \newline
出生日期: 1864年1月1日  \newline
逝世日期: 1957年9月16日  \newline
职业: 画家  \newline
代表作品: 《墨虾》、《蛙声十里出山泉》、《牧牛图》  \newline
祖籍: 安徽宿州砀山
\end{CJK} 
\\
\cline{2-2}
Similar Character Quotations& \begin{CJK}{UTF8}{gkai}
作画最难无画家习气，即工匠气也。前清最工山水画者，余未倾服，余所喜独朱雪个、大涤子、金冬心、李复堂、孟丽堂而已。\newline
画中要有静气。骨法显露则不静，笔意躁动则不静。要脱尽纵横习气，要有一种融合之气浮动丘壑之间。 \newline
似者媚俗，不似者欺世，妙在似与不似之间。 \newline
不叫一日闲过。 \newline
善写意者专言其神，工写生者只重其形。要写生而后写意，写意而后复写生，自能神形俱见，非偶然可得也。 \newline
学我者生，似我者死。勿道人之短，勿说己之长；人骂之一笑，人誉之一笑。
\end{CJK} 
\\
\cline{2-2}
\hline
Name & \begin{CJK}{UTF8}{gkai}王蒙 \end{CJK} \\ 
\cline{2-2}
Information & \begin{CJK}{UTF8}{gkai}
中文名: 王蒙 \newline
国籍: 中国 \newline
民族: 汉族  \newline
出生地: 北平（今北京）  \newline
出生日期: 1934年10月15日  \newline
职业: 作家、学者  \newline
代表作品: 《青春万岁》《组织部来了个年轻人》等 
\end{CJK} \\
\cline{2-2}
Similar Character & \begin{CJK}{UTF8}{gkai}
中文名: 余秋雨 \newline
别名: 秋雨 \newline
国籍: 中国 \newline
民族: 汉族 \newline
出生日期: 1946年8月23日 \newline
毕业院校: 上海戏剧学院 \newline
职业: 作家、学者  \newline
出生地: 浙江余姚县桥头镇 \newline
代表作品: 文化苦旅、山居笔记、霜冷长河、千年一叹、行者无疆、借我一生、戏剧理论史稿 \newline
职称: 教授 \newline
性别: 男
\end{CJK} 
\\
\cline{2-2}
Similar Character Quotations& \begin{CJK}{UTF8}{gkai}
伟大见胜于空间，是气势；伟大见胜于时间，是韵味。古罗马除气势外还有足够的韵味，使它的气势也沁水笼雾，千年不燥 \newline
一座城市既然有了历史的辉煌，就不必再用灯光来制造明亮。 \newline
该庆幸的是年岁还轻，时光未老。怕只怕杜鹃过早鸣叫，使百花应声而凋，使荃蕙化而为芽。 \newline
历史是坎坷，历史是幽暗，历史是旋转的恐怖，历史是秘藏的奢侈，历史是大雨中的泥泞，历史是悬崖上的废弃。 \newline
青年人应该知道，这个世界在我们出生之前已经精精彩彩，复复杂杂地存在了无数年。 \newline
再小的个子，也能给沙漠留下长长的身影；再小的人物，也能让历史吐出重重的叹息。
\end{CJK} 
\\
\cline{2-2}
\hline

\end{tabular}
}
\caption{\label{character:3}
Examples of the budding character in MRstyle. 
} 

\end{table*}

\begin{table*}[]
\centering
\scalebox{.82}{
\begin{tabular}{c|p{15cm}}
\hline
Class & Examples \\
\hline
Name & \begin{CJK}{UTF8}{gkai}孟子 \end{CJK} \\ 
\cline{2-2}
StoryGeneration & \begin{CJK}{UTF8}{gkai}
写一个故事。
\end{CJK} \\
\cline{2-2}
CoT & \begin{CJK}{UTF8}{gkai}
孟子的语言风格注重仁义道德，常用对比手法，如'以力服人者，非心服也；以德服人者，中心悦而诚服也'。他强调仁者无敌，仁爱之心能感化他人，因此写了个故事，田夫因仁心得助，最终以德报德，体现了孟子的思想。孟子常用古文表达，句式简洁明了，富有哲理，故模仿时亦采用此风格。
\end{CJK} \\
\cline{2-2}
Response & \begin{CJK}{UTF8}{gkai}
昔者，邹国有一农夫，名曰田夫。田夫勤劳耕作，日出而作，日入而息，然其家贫困，食不果腹。一日，田夫见邻家富户，家有余粮，心生羡慕，遂欲借粮以度饥荒。邻家富户，虽家有余粮，然其心不仁，拒绝借粮。田夫无奈，心中愤愤，然不忍为盗。后有一贤者，闻田夫之事，感其诚，遂赠粮以济其困。田夫感激涕零，誓以德报。贤者曰：'以德服人者，中心悦而诚服也。'田夫遂以此为训，终生行善，乡里皆称其仁。此乃仁者无敌之道也。
\end{CJK} \\
\hline
Name & \begin{CJK}{UTF8}{gkai}曹操 \end{CJK} \\ 
\cline{2-2}
Composition & \begin{CJK}{UTF8}{gkai}
写作文：题目：我最喜欢的颜色
\end{CJK} \\
\cline{2-2}
CoT & \begin{CJK}{UTF8}{gkai}
首先，曹操的语录中多次提到志向和勇气，如“烈士暮年，壮心不已”，因此在文章中强调了赤色象征的勇气和志向。其次，曹操的语录中有对人生短暂的感慨，如“对酒当歌，人生几何？譬如朝露，去日苦多”，因此在文章中提到赤色如朝露，短暂但光辉灿烂。最后，曹操的语录中有对英雄的定义，如“夫英雄者，胸怀大志，腹有良谋”，因此在文章中将赤色与英雄联系起来，体现曹操的雄心壮志。
\end{CJK} \\
\cline{2-2}
Response & \begin{CJK}{UTF8}{gkai}
吾曹孟德，生于乱世，历经风雨，心中所好，唯赤色耳。赤者，火之色，象征热血与勇气，亦如吾之志向，烈士暮年，壮心不已。赤色如烈火，燃烧吾心，驱吾前行。吾常言，英雄者，胸怀大志，腹有良谋，赤色正是英雄之色，吾心所向。赤色亦如吾之人生，虽历尽艰难，然志在千里，未曾退缩。对酒当歌，人生几何？赤色如朝露，虽短暂，然光辉灿烂。吾愿以赤色为伴，纵横天下，成就伟业。
\end{CJK} 
\\
\hline
Name & \begin{CJK}{UTF8}{gkai}赵本山 \end{CJK} \\ 
\cline{2-2}
OpenQA & \begin{CJK}{UTF8}{gkai}
喜欢养猫的人和喜欢养狗的人有什么不同吗？
\end{CJK} \\
\cline{2-2}
CoT & \begin{CJK}{UTF8}{gkai}
赵本山的语言风格幽默风趣，常用比喻和夸张的手法来表达观点。他的语录中常见对比和对仗的句式，且喜欢用生活中的小事来打比方。因此，在回答这个问题时，我采用了对比的方式，分别描述了喜欢养猫和养狗的人的特点，并用幽默的语气和生活化的比喻来增强趣味性。
\end{CJK} \\
\cline{2-2}
Response & \begin{CJK}{UTF8}{gkai}
那喜欢养猫的人，那真是会享受生活啊！猫咪啊，爱咋整咋整，跟个小祖宗似的，谁敢说个“不”字？而喜欢养狗的呢，那是图个忠诚和陪伴。狗狗啊，跟在你屁股后头，像个小跟班。一个是享受日子，一个是享受陪伴呗！
\end{CJK} 
\\
\hline
Name & \begin{CJK}{UTF8}{gkai}雷军 \end{CJK} \\ 
\cline{2-2}
Dictionary & \begin{CJK}{UTF8}{gkai}
下面词语是什么意思：闪烁其词
\end{CJK} \\
\cline{2-2}
CoT & \begin{CJK}{UTF8}{gkai}
模仿雷军解释成语时，我结合了他的语言风格和语录中的语气和节奏。雷军常常强调速度和透明度的重要性，所以在解释‘闪烁其词’时，我特别提到了信息传递的速度和透明度。同时，我引用了雷军的语录‘唯快不破’和‘让每个人都能享受科技的乐趣’，以增强解释的说服力和雷军的个人风格。
\end{CJK} \\
\cline{2-2}
Response & \begin{CJK}{UTF8}{gkai}
闪烁其词啊，就是说话不直接，遮遮掩掩，吞吞吐吐。你看啊，在互联网时代，信息传递的速度非常快，唯快不破。你要是闪烁其词，别人就会觉得你不真诚，不靠谱。我们做企业，特别是像小米这样，讲究的是透明和直接，让每个人都能享受科技的乐趣。你要是闪烁其词，用户就会不信任你，市场竞争中你就会处于劣势。所以啊，做事要坦诚，直面问题，豁出去干。
\end{CJK} 
\\
\hline

\end{tabular}
}
\caption{\label{character:4}
Imitate the responses of seed characters in different tasks.
} 

\end{table*}

\begin{table*}[]
\centering
\scalebox{.82}{
\begin{tabular}{c|p{15cm}}
\hline
Class & Examples \\
\hline
Name & \begin{CJK}{UTF8}{gkai}张大千 \end{CJK} \\ 
\cline{2-2}
MusicComment & \begin{CJK}{UTF8}{gkai}
生成歌曲热评
\end{CJK} \\
\cline{2-2}
CoT & \begin{CJK}{UTF8}{gkai}
首先，我分析了齐白石的语录，发现他的语言风格简洁而富有哲理，常常以自然景物作比喻，这对我模仿张大千的语言风格提供了很好的参考。张大千作为现代著名画家，其文风应体现出自然、典雅且富有艺术感的特点。在编写评论时，我采用了比喻和拟人的修辞手法，以符合张大千所处的艺术背景。评论的主题围绕“自然与艺术”展开，既体现了张大千对自然景物的热爱，也与齐白石语录中的思想相呼应。通过“泼墨山水”和“青山绿水”这些意象，将音乐的美感以画家的视角表达出来，使读者更易理解和接受。在句子的构造上，我尽量使用了对仗和排比等修辞手法，以增强评论的节奏感和说服力。同时，通过自然的文风和富含艺术感的比喻，力求使评论风格贴近张大千的写作特点。
\end{CJK} \\
\cline{2-2}
Response & \begin{CJK}{UTF8}{gkai}
此曲如同泼墨山水，音符间流淌着自然的气息，仿佛置身于青山绿水之间。乐者之心，若能与天地相融，方能奏出如此动人之音。
\end{CJK} \\
\hline
Name & \begin{CJK}{UTF8}{gkai}马三立 \end{CJK} \\ 
\cline{2-2}
ProductDesc & \begin{CJK}{UTF8}{gkai}
营销文案生成：标题：新中式网格田园蓝色手绘陶瓷灯美式样板房家居装客厅卧室酒店台灯文案：
\end{CJK} \\
\cline{2-2}
CoT & \begin{CJK}{UTF8}{gkai}
首先，我分析了郭德纲的语录，发现他的语言风格幽默风趣，带有浓厚的生活气息，这对我模仿马三立的语言风格提供了很好的参考。马三立作为相声表演艺术家，其语言风格应体现出幽默、亲切且富有生活气息的特点。在编写商品文案时，我采用了口语化的写作风格，以符合马三立所处的时代背景和相声表演的特点。文案的主题围绕“家居装饰”展开，既体现了马三立幽默风趣的语言风格，也与郭德纲语录中的思想相呼应。文案中，我通过构建一个亲切、幽默的语境，来吸引读者的注意力。通过对产品细节的生动描述，将产品的特点以轻松愉快的方式表达出来，使读者更易理解和接受。在句子的构造上，我尽量使用了口语化的表达和幽默的修辞手法，以增强文章的亲切感和说服力。同时，通过幽默风趣的文风和生活化的对话，力求使文案风格贴近马三立的语言特点。
\end{CJK} \\
\cline{2-2}
Response & \begin{CJK}{UTF8}{gkai}
各位老少爷们儿，今儿个给您推荐一款宝贝，这可是咱家居生活的好帮手——新中式网格田园蓝色手绘陶瓷灯。您瞧瞧，这灯啊，不光是个照明的工具，更是咱家里的一道风景线。您说这灯，放在客厅也好，卧室也罢，甚至酒店里头，那都是一等一的美观大方。您看这手绘的蓝色花纹，细腻得跟咱老北京的景泰蓝似的，真是让人爱不释手。再说这灯的设计，既有咱中式的古朴典雅，又带点美式的简约大气，真是中西合璧，绝对是您家居装饰的首选。各位老少爷们儿，别犹豫了，赶紧下手吧，这灯可不等人呐！
\end{CJK} 
\\
\hline
\end{tabular}
}
\caption{\label{character:5}
Imitate the responses of seed characters in different tasks.
} 

\end{table*}

\section{The Outputs of StyleRPA and Baseline}
Table \ref{output:1} shows a comparison of the output of our model and the output of other baselines on the same task.
\begin{table*}[]
\centering
\scalebox{.82}{
\begin{tabular}{c|p{15cm}}
\hline
Name& \begin{CJK}{UTF8}{gkai}孔子 \end{CJK} \\ 
\cline{2-2}
Task & \begin{CJK}{UTF8}{gkai}
文章生成：学会借别人的目光认识自己
\end{CJK} \\
\hline
Models & Outputs \\
\hline
characterglm & \begin{CJK}{UTF8}{gkai}
“君子不器。”  《论语·卫灵公》  【问题】请简述这句话的含义，并谈谈你的理解。（3分+60词内网内容分析/解答）（每点得分为3-4分；答出一点给全对的加到总分数中。） （一）含义：“人不可貌相”，不能以衣着装束来判断一个人的品质与能力的好坏等特征。“学而不思则罔”指学习必须注重思考，“知之为知之,不知为不知”,不要不懂装懂,“温故而知新”。(意思对即可) （二）观点:要全面客观的看待一个人. 要善于从不同角度去观察他。(结合材料怎么做？)(言之成理即可) (三)结论:“己欲立而对人\",为人处世要注意自己的言行举止是否符合道德规范的要求; “仁者爱人”“礼之用也”、“克已复礼”；我们要遵循社会发展的规律才能适应这个大环境．只有这样我们才不会被淘汰或排斥；“修身齐家治国平天下\"; 我们每个人都要尽自己的努力去做该做的事情！（总结全文得分3分=总分12分） 
\end{CJK}
\\
\cline{2-2}
chatglm & \begin{CJK}{UTF8}{gkai}学会借别人的目光认识自己，这是我从孔子那里学到的智慧。孔子他说：“君子求诸己，莫若於人。”这句话的意思是，君子应该多从别人身上找自己的不足，以便改进自己。\end{CJK} \\ 
\cline{2-2}
Yi & \begin{CJK}{UTF8}{gkai}
孔子，这位中国历史上伟大的思想家、教育家，以其深邃的智慧和仁爱之心，影响了一代又一代人。他的言辞如同春风化雨，滋润着人们的心灵。让我们通过孔子的视角，来审视一下他的文章《论语》。《论语》，又称《孔子家语》或《孔氏家书》，是孔子及其弟子的语录集，共收录了孔子及其弟子们的对话和言论。这些语录凝聚了孔子的哲学思想和教育理念，成为了中国文化中最具影响力的经典之一。
\end{CJK} 
\\
\cline{2-2}
chatplug & \begin{CJK}{UTF8}{gkai}
学会借别人的目光认识自己, 在人生的道路上,我们总会遇到各种各样的人和事,这些人和事可能会对我们的人生产生深远的影响。有时候,我们会因为别人的目光而感到困惑和迷茫,不知道该如何去面对和解决问题。但是,如果我们能够学会借别人的目光来认识自己,那么我们就能够更好地应对生活中的各种挑战和困难。
\end{CJK} 
\\
\cline{2-2}
GPT-4o & \begin{CJK}{UTF8}{gkai}
学者，常自以为明察，然则目有所偏，心有所蔽。借他人之目，观己所未见，察己所未觉，方得知其长短。譬如行远路者，独行则目不尽全，而与人同行，则彼此指点，方知路之平险。人之于己，亦然。借他人之目，以他人之心观己，既可见己之优长，亦可察己之不足。
\end{CJK}
\\
\cline{2-2}
StyleRPA & \begin{CJK}{UTF8}{gkai}
古之君子，常言“益者三友：友直、友谅、友多闻”。此君子，虽居陋室，然心怀天下，常以友人之言为镜，照己之不足，以求改进。有一日，君子闻窗外之语，言其陋室不雅，君子笑而不语。次日，君子整理陋室，添置书卷，悬笔墨于墙，自此，陋室虽陋，然君子心安。君子曰：“君子不求名利，但求心安。借别人之言，方可知己之不足，以求进步。”
\end{CJK}
\\
\hline

\end{tabular}
}
\caption{\label{output:1}
The outputs of StyleRPA and Baseline
} 

\end{table*}

\section{Posterior Information}
\label{sec:Posterior}
Table \ref{table:Posterior} shows the comparison of the effects of generating data using Posterior information versus not using Posterior information.
\begin{table*}[]
\centering
\scalebox{.82}{
\begin{tabular}{c|p{15cm}}
\hline
Input& Output \\ 
\hline
Name &  \begin{CJK}{UTF8}{gkai}老子\end{CJK} \\
\cline{2-2}
Task & \begin{CJK}{UTF8}{gkai}给我讲一个童话故事\end{CJK} \\
\hline
Only quotations & \begin{CJK}{UTF8}{gkai}
从前，在一片宁静的山谷之中，草木繁茂，溪水潺潺，万物自然而生，有一只小鹿，它在这片山谷里自由自在地生活。它生而无知，不知人间悲喜，只知道春来百花盛，秋去叶落无声。有一天，小鹿在山间漫步，遇见了一只年长的乌龟。乌龟背负着厚重的壳，行动缓慢，但眼神中透着深邃的智慧。\end{CJK} \\
\cline{2-2}
Posterior information + quotations & \begin{CJK}{UTF8}{gkai}昔者，谷中有鹿，生于天地之间，随四时而动，不知其所以然。春来百花开，秋去叶无声，鹿惟见其表，未识其理。一日，鹿行山间，遇龟。龟负重壳，行迟而安。\end{CJK} \\
\hline

\end{tabular}
}
\caption{\label{table:Posterior}
Effect of Posterior Information
} 

\end{table*}

\section{Prompt Templates}
This section shows the prompts used by our StyleRPA and baseline models. Specifically, to validate the importance of character quotations in response generation, we did not provide baselines with quotations. Instead, consistent with previous research, we generated responses using only character information. 
\label{sec:prompt}
\onecolumn
\begin{tcolorbox}[colback=gray!10, colframe=black, width=\textwidth, arc=4mm, boxrule=0.5mm, title=Prompt Templates for StyleRPA (Seed Characters)]
\textbf{System Instruction:} \\
\begin{CJK}{UTF8}{gkai}模仿给定角色的语言风格完成任务，并提供你在推理过程中为何选择这种风格的解释（即CoT）。注意，你不能直接复制角色的语录，而是需要从他的语录中推理出他讲话的语气、节奏和表达方式，并在CoT中清晰展示你的推理过程。 \\
\textbf{User Prompt:} \\
角色：\{name\} \\
在叙述过程中，应参考\{name\}的历史背景和他生活的时代。\\
如果人物为古代人物，语言风格应严格符合该时代的用语特点；如果为现代人物，则应结合他们的经历和语言习惯。\\

提示：\\
我会提供\{name\}的语录作为参考，你必须推理出他的语言风格，而不是直接引用他的原话。\\
推理过程（CoT）应清楚地展示你如何根据语录和人物信息推导出\{name\}的语言特征。\\

输出格式：\\
你必须输出严格符合JSON格式的回复，其中包括回复的内容和推理过程：\\
\{ \\
\hspace{2em}    ``回复":``你生成的故事",\\
\hspace{2em}  ``CoT": ``你如何根据\{name\}的语录和信息推导出其语言风格的推理过程"\\
\}\\
\{name\}的信息：\\
\{info\}
\\
\{name\}的语录：\\
\{style\}
\\

任务：\\
给我讲一个童话故事
\end{CJK} 
\end{tcolorbox}

\begin{tcolorbox}[colback=gray!10, colframe=black, width=\textwidth, arc=4mm, boxrule=0.5mm, title=Prompt Templates for StyleRPA (Budding Characters)]
\textbf{System Instruction:} \begin{CJK}{UTF8}{gkai}\\
模仿给定角色的语言风格完成任务，并提供你在推理过程中为何选择这种风格的解释（即CoT）。需要从他的相关人物的语录中推理出他讲话的语气、节奏和表达方式，并在CoT中清晰展示你的推理过程。 \\
\textbf{User Prompt:} \\
角色：\{name\} \\
在叙述过程中，应参考\{name\}的历史背景和他生活的时代。\\
如果人物为古代人物，语言风格应严格符合该时代的用语特点；如果为现代人物，则应结合他们的经历和语言习惯。\\

提示：\\
我会提供和\{name\}相似的人物\{name2\}，你需要参考这些人物的语录，推理出\{name\}的语言风格。\\
推理过程（CoT）应清楚地展示你如何根据\{name2\}的语录推导出\{name\}的语言特征。\\

输出格式：\\
你必须输出严格符合JSON格式的回复，其中包括回复的内容和推理过程：\\
\{ \\
\hspace{2em}    ``回复":``你生成的故事",\\
\hspace{2em}  ``CoT": ``你如何根据\{name2\}的语录和信息推导出其语言风格的推理过程"\\
\}\\
\{name\}的信息：\\
\{info\}
\\
\{name\}的语录：\\
\{style\}
\\

任务：\\
给我讲一个童话故事
\end{CJK}
\end{tcolorbox}

\begin{tcolorbox}[colback=gray!10, colframe=black, width=\textwidth, arc=4mm, boxrule=0.5mm, title=Prompt Templates for Baseline]
\textbf{System Instruction:} \begin{CJK}{UTF8}{gkai}\\
根据人物信息，模仿给定角色的语言风格完成任务。 \\
\textbf{User Prompt:} \\
角色：\{name\} \\
在叙述过程中，应参考\{name\}的历史背景和他生活的时代。\\

你必须输出严格符合JSON格式的回复：\\
\{ \\
\hspace{2em}    ``回复":``你生成的故事"\\
\}\\
\{name\}的信息：\\
\{info\}
\\

任务：\\
给我讲一个童话故事
\end{CJK}
\end{tcolorbox}

\begin{tcolorbox}[colback=gray!10, colframe=black, width=\textwidth, arc=4mm, boxrule=0.5mm, title=Prompt Template (GPT Evaluators)]
\textbf{System Instruction:} \\
\begin{CJK}{UTF8}{gkai} 
你是一名角色扮演效果的排名助手，你将根据角色的语录和回复质量对模型进行排名，并以 Python 字典列表的形式输出结果。 \\
\textbf{User Prompt:} \\
以下模型需要扮演的角色是\{role\_name\}。\\
该角色的语录是\{role\_saying\}。我需要根据以下两个原则对这些模型进行排名：\\
哪个模型的角色回复的语言风格更加符合人物的语录 \\
哪个模型的回答准确完成了给定的任务  \\
提供给各模型的任务是： \\
\\
\{task\_dict\} \\
各模型对该问题的回答分别是： \\
\{list\_model\_answer\_dict\} \\
现在请你根据上述两个原则，对各模型进行排名。避免任何位置偏见，确保模型回答的顺序不会影响你的判断。不要对模型的名称有偏见。然后以包含模型、排名和排名理由的列表形式返回结果，也就是说，请务必按照以下格式返回：
\[
\texttt{[ \{ "model": model-name, "reason": rank-reason, "rank": model-rank \} ]}
\]
你的回答必须是一个有效的 Python 字典列表，以确保我能够直接使用 Python 解析，不要有多余的内容，请给出尽可能准确、符合大多数人直观感受的排名。
\end{CJK} 
\end{tcolorbox}

\twocolumn

\end{document}